\documentclass{article}

\usepackage[final, nonatbib]{nips_2018}

\usepackage[utf8]{inputenc} 
\usepackage[T1]{fontenc}    
\usepackage{hyperref}       
\usepackage{url}            
\usepackage{booktabs}       
\usepackage{amsfonts}       
\usepackage{nicefrac}       
\usepackage{microtype}      

\title{Dynamic Scheduling of MPI-based Distributed Deep Learning Training Jobs}

\author{
  Tim Capes \\
  SAIC-Toronto\\
  \texttt{t.capes@samsung.com} \\
  \And
  Vishal Raheja \\
  SAIC-Toronto \\
  \texttt{v.raheja@samsung.com}
  \AND
  Mete Kemertas \\
  SAIC-Toronto \\
  \texttt{mete.kemertas@samsung.com}
  \And
  Iqbal Mohomed \\
  SAIC-Toronto \\
  \texttt{i.mohomed@samsung.com}
}

\begin{document}

\maketitle
\begin{abstract}
 There is a general trend towards solving problems suited to deep learning with more complex deep learning architectures trained on larger training sets. This requires longer compute times and greater data parallelization or model parallelization. Both data and model parallelism have been historically faster in parameter server architectures, but data parallelism is starting to be faster in ring architectures due to algorithmic improvements. In this paper, we analyze the math behind ring architectures and make an informed adaptation of dynamic scheduling to ring architectures. To do so, we formulate a non-convex, non-linear, NP-hard integer programming problem and a new efficient doubling heuristic for its solution. We build upon Horovod: an open source ring architecture framework over TensorFlow.  We show that Horovod jobs have a low cost to stop and restart and that stopping and restarting ring architecture jobs  leads to faster completion times. These two facts make dynamic scheduling of ring architecture jobs feasible.  Lastly, we simulate a scheduler using these runs and show a more than halving of average job time on some workload patterns.
\end{abstract}

\section{Introduction} \label{sec:introduction}

As deep learning applications become more pervasive, organizations such as research labs, have continued to make large investments in GPU clusters. A major challenge that follows is the need to better utilize cluster resources. In our organization, we have observed scientists using GPUs on a single server due to the complexities of multi-server training. Even when a scientist writes code to leverage multiple GPUs, they face the problem of sharing the cluster effectively in the face of contention.

In this paper, we argue that \textit{dynamic scaling of the number of GPUs available to deep learning tasks is critical to effective utilization of clusters}. While others \cite{Optimus} have made this argument in relation to deep learning training via parameter servers, we explore the issue with training over ring architectures - an emerging approach for DL training that offers better resource utilization for data parallel workloads.

In this paper, we start by providing a background and motivation for ring architecture based DL training (Section 2). We then derive the mathematics behind the communications overhead. We investigate the difference for having a number of workers equal to a power of 2 or not (Section 3). We next formulate the scheduling problem and a new doubling heuristic for its solution.  We outline why this doubling heuristic respects the aforementioned communication overheads and enables large allocations when other heuristics would fail to do so (Section 4). We continue by providing experimental results on dynamic resizing with a ring architecture approach, and show both that scaling is effective at improving completion time of DL training tasks, and that the stop-restart times are negligible (Sections 5 \& 6). Based on experimental runs, we simulate a scheduler yielding improved cluster performance (Section 7).

\section{Background} \label{sec:background}

While cluster schedulers have a long history in academia and practice, system support for distributed DL training in schedulers is in its infancy. One useful capability is dynamic scheduling, where the resources allocated to a particular training  job are updated based on factors such as job priority, contention, resource availability, and overall demand.

Optimus \cite{Optimus} is a dynamic scheduler for parameter server deep learning jobs written for MXNet.  Optimus uses online fitting to model the number of epochs till convergence.  It uses resource modelling with non-negative least squares (NNLS) to learn the speed per epoch for $p$ parameter servers and $w$ workers.  These are used to formulate a non-linear non-convex NP-hard integer programming problem. This problem is heuristically solved with the greedy heuristic of adding the best projected worker or parameter server at each step. Our work extends Optimus by enabling it to work on ring architectures using a novel doubling heuristic.

TensorFlow was specifically designed to be able to move away from parameter servers \cite{TensorFlow}.  One architecture benefiting from this open design is a ring architecture: an architecture where all nodes are workers and workers use collective communication for parameter updates. Horovod\cite{Horovod}, uses both NCCL (Nvidia Collective Communications Library) \cite{NCCL} and OpenMPI \cite{OpenMPI} to provide a ring architecture implementation over Tensorflow. In the paper, "ImageNet in 1 hour" researchers from Facebook showed the community a way to vastly improve GPU utilization\cite{ImageNet1Hour}.  Horovod had similar improvements for TensorFlow \cite{Horovod}.

A major portion of the efficiency gains above is due to increasing use of High Performance Computing (HPC) methods in OpenMPI and NCCL for deep learning. NCCL is a collective communications library that enables some efficient algorithms at the GPU level, notably \textit{all reduce}; which is an efficient way to share loss and gradient vectors across multiple GPUs \cite{NCCL}.  OpenMPI enables similar primitives at the cross-server level \cite{OpenMPI}.  Horovod uses both NCCL and OpenMPI for \textit{all reduce} based training of neural network models \cite{Horovod}.  Techniques utilizing \textit{all reduce} have scaled to $512$ GPUs and substantially larger minibatch sizes than those using parameter servers \cite{Horovod} \cite{ImageNet1Hour}.

\subsection{Algorithms for ring architectures}
Systems leveraging ring architectures are heavily influenced by \cite{Rabenseifner} which introduced the doubling halving algorithm. This key algorithm enables a latency efficient all-to-all copy and sum (\textit{all reduce}). We describe the implementations in this section with a focus on doubling-halving which is used by our doubling heuristic.

Let $w$ be the number of GPU's used by a single job and let $n$ be the size of the gradient tensors being exchanged. The ring algorithm divides the gradient tensors into $w$ segments and copies a segment to each GPU taking $w-1$ steps and copying $\frac{n}{w}$ data per GPU at each step \cite{Rabenseifner}. This algorithm is efficient for very large data sizes in deep learning. However, it has a tradeoff of having latency that is linear in the number of GPUs.  For parameter sizes up to $10^7$, the doubling-halving algorithm for powers of $2$ has been found to be significantly more efficient \cite{Horovod}.

The doubling-halving algorithm \cite{Rabenseifner} has $\log(w)$ steps and at step $i$, node $j$ would copy the forward or backward half of it's current vector to node $j+2^{i-1}$. For example, at step $1$, odd nodes copy the forward half and even nodes copy the backward half.  At the end of this halving, each node ends up with all of the data for a segment of length $\frac{n}{w}$. The segments are then combined by taking $\log(w)$ steps in reverse, starting by copying at distance $w/2$ and finishing by copying at distance $1$ with each copy doubling in size. This algorithm has low-latency due to fewer requests. However, this only works for powers of 2, otherwise, a variation known as binary blocks is used.

The binary blocks algorithm \cite{Rabenseifner} works in a manner similar to the doubling-halving algorithm but with modifications based on building up blocks of size power of 2. If the number of GPUs is a power of 2 it's identical, otherwise, it represents $w$ as a sum of powers of $2$ and executes a similar strategy with some extra steps to aggregate the inexact matches.  The binary blocks algorithm is most efficient when there are small differences between the powers of $2$ in the sum and does worse when the powers of $2$ in the sum have large differences \cite{Rabenseifner}. 

\section{Performance modelling of ring-reduce deep learning jobs} \label{sec:performance}

As in Optimus\cite{Optimus}, we propose performance modelling machine learning jobs by a two-step process: first, model the number of epochs through online fitting, and second, model time per epoch for given resource configurations. In section 3.1 we derive the first estimate, the number of epochs to convergence. In section 3.2, we derive the second estimate $f(w)$ the velocity in epochs per second. These two estimates together allow the runtime of a job to be estimated.

\subsection{Online learning of convergence in epochs}
Let $k$ be batch step and $l$ be loss and $\beta_i$ the parameters to be solved for. Since SGD converges at O($1/k$) we fit the following equation using NNLS with $\beta_0>0$:
\begin{equation}
l = \frac{1}{\beta_0*k+\beta_1} +\beta_2
\end{equation}
\subsection{Resource modelling}

We next propose a resource-to-speed model that accounts for all the different algorithms used in ring architectures for \textit{all reduce}.

We now derive time taken per minibatch as a combination of: forward propagation time, \textit{all reduce time}, backward propagation time and communication overhead. Suppose there are $w$ workers on the job.  We model the forward propogation time as the size of a minibatch ($m$) mutiplied by the time to process an example:  $m* T_{forward}$. We use $T_{back}$ to represent the back propogation of a single example and thus model backpropagation of a minibatch as $m*T_{back}$.  We rely on data parallelism in TensorFlow and Horovod to distribute this computation across the workers and use gradient exchange to update. We model \textit{all reduce} in accordance with each algorithm, where $\alpha$ is the latency per message, $\beta$ is the transfer time per byte, $\gamma$ is the computation cost per vector byte, and $n$ is the size of the model. $T_{ring}$ is the time for ring architectures with the ring algorithm, $T_{dh}$ is the time for ring architectures with doubling-halving and $T_{bb}$ is the time for ring architectures with binary blocks. The coefficients of $\alpha$, $\beta$ and $\gamma$ are based on the underlying collective communications primitives \cite{MPICH}.

\begin{equation}
T_{ring} = m*(T_{forward}+T_{back}) + (w-1) * 4 * \alpha + (w-1)*(n/w)*4*\beta +(w-1)*(n/w)*2*\gamma
\end{equation}
\begin{equation}
T_{dh} = m*(T_{forward}+T_{back})  + 4* \log(w) * \alpha + 4*n*\beta + \frac{5}{2}*n*\gamma
\end{equation}
\begin{equation}
T_{bb} = m*(T_{forward}+T_{back}) + (5+4*\lceil\log(w)\rceil)*\alpha + 7*n*\beta + 3*n*\gamma
\end{equation}
\begin{equation} f(w) = (\theta_0 * (m/w) + \theta_1*(w-1) + \theta_2 * (w-1)*(n/w)+ \theta_3)^{-1}
\end{equation}

$f(w)$ is our resource model where $\theta$'s are positive coefficients to be learned for each job.  It's beneficial to introduce $\theta_3$ to collect constant terms. The same $f$ can be used to model all three $T$'s but the behaviours of the coefficients will be different. This different behaviour has consequences for the heuristic used to schedule. \textit{In all cases, we can fit a NNLS model for each value of w}.   To learn the values of $\theta$'s, we collect data points of the form $(w,f(w))$. 
\section{Dynamic scheduling}
In a typical DL cluster, jobs arrive in an online manner.  We model a periodic allocation of resources to active jobs by adjusting the number and placement of workers for each job in the shared DL cluster. Before scheduling a job, the scheduler must train its model on a small set of training data for several steps with possible values of $w$ (precompute) or it must execute a data gathering strategy so that it can subsequently schedule efficiently (explore).
\subsection{Resource Allocation}
In each scheduling interval, let $Q_j$ be the remaining epochs to achieve model convergence as predicted in section 3.1. $f(w_j)$ is the training speed for job $j$ with $w_j$ workers in epochs per second as predicted in section 3.2. Let $C$ be the GPU capacity and let $t_j$ be the time taken for job j. We model the scheduling problem as: \newline
Minimize: $$\sum_{j \in J} t_j$$
Subject to:$$t_j = \frac{Q_j}{f(w_j)}, \large \forall j \in J$$
$$\sum_{j \in J} w_j \leq C $$
$$w_j \in Z^{+},\large \forall j \in J$$

\subsection{Heuristic job assignment}

Our doubling heuristic works by assigning 1 worker to every job, then assigning additional workers by doubling based on the average marginal gain:

\begin{equation}(\frac{Q_j}{f(w_j)} - \frac{Q_j}{f(2*w_{j})})/w_{j} \end{equation}

The assignments, $w_j$, occur by picking the maximum of these time improvements at each step and assigning an additional $w_j$ workers to the task that maximizes per GPU gain.  This doubling heuristic is important as some increases in GPUs have worse algorithmic performance leading to the Optimus heuristic getting stuck in local optima. For example, going from 8 to 9 GPUs will have worse per GPU performance than going from 8 to 16 GPUs. This means that even though going to 16 GPUs could be desirable it may not be reachable without the heuristic change because the poor score for going from 8 to 9 GPUs would block exploring 16 GPUs.  The changes are also useful for practical aspects as limiting possible assignments enables more precomputation. Simulating every run to an adequate level of accuracy requires tens of seconds per configuration and limiting the available configurations dramatically improves simulation time.

\subsection{Task placement}
Task placement is the allocation of jobs to specific hardware once the allocation sizes have been solved by the scheduler.  Task placement is much simpler for ring architectures as there are no parameter servers.  It is still important to allocate as few total nodes as possible for the same number of GPUs. The task placement portion of scheduling can be solved straightforwardly by standard algorithms and the innovations of this paper are in scheduling not placement.

\section{Experimental setup}

We choose the ResNet model for image classification \cite{ResNetv2} as our primary deep learning model and task. ResNet has achieved widespread adoption across a variety of tasks and datasets, and is often used as a feature extractor for downstream tasks. We choose not to use the bottleneck version of the model, therefore, our network depth is required to be 6n+2 \cite{Kaiming}. For realism, we choose a relatively deeper model size of 110. For training time considerations, we run our experiments on the CIFAR-10 dataset \cite{CIFAR10}, which contains more than an order of magnitude fewer images than ImageNet.

Our experimental hardware consists of a single node having 8 Tesla K40m GPUs with an active memory of \textasciitilde11.25 GB each, 40 Intel Xeon E5-2670 cores with a clock rate of 2.50GHz and an Infiniband interconnect with 100Gbit/sec (4x EDR) throughput. In our setup, the number of cores that will be used is going to be equal to the number of GPUs being used for a given training job. We leverage the official TensorFlow implementation of ResNet\footnote{https://github.com/tensorflow/models/tree/master/official/resnet}, and use Tensorflow v1.8.0, CUDA v9.0, cuDNN v7.0.5, OpenMPI v2.1.3, NCCL v2.1.15 and Horovod v0.13.11.

As baselines, we first run the Resnet training on 4 GPUs and 8 GPUs without any resizing. We then rerun the training job initially with 4 GPUs, checkpoint and stop at different steps during training and restart the training job with 8 GPUs. While resizing the configuration, we keep the number of examples per GPU constant at 128. So the global batch size increases with the increase in the number of GPUs. We adjust the initial learning rates according to the suggestion given in \cite{ImageNet1Hour}:

\begin{equation}\label{learning_rate_eqn}
 lr_{new} = \frac{\#GPUs_{new}}{\#GPUs_{last}} \times lr_{last}
\end{equation}

We ensure the learning decay occurs at the appropriate number of epochs with the change in the global batch sizes. The suggested initial learning rate for 1 GPU with batch size of 128 is $0.1$. So, we take the initial learning rates for 4 GPUs as $0.4$ and 8 GPUs as $0.8$. We divide the learning rate by $10$ at epochs $100$ and $150$.

\section{Experimental results}

In this section, we show that checkpointing is time efficient. That is to say start and stop time is negligible and convergence of jobs from checkpointing and resuming with additional hardware is rapidly accelerated. We run some experiments with a Horovod job that is initially scheduled on 4 GPUs and restores from checkpoint to run on 8 GPUs.

\tabcolsep=0.11cm
\begin{table}[]
\caption{Profiling results on ResNet-110 for a minibatch size of 128 images per GPU.}
\centering
\begin{tabular}{|c|c|c|c|c|}
\hline
\# GPUs & $T_{forward}$ (ms) & $T_{back}$ (ms) & $T_{total}$ (ms) & images/sec \\ \hline
1       & 108.0              & 236.5          & 402.5       & 318.0      \\ \hline
2		& 110.2				& 274.6		& 427.2			& 576.2    \\ \hline
4       & 107.1              & 290.1          & 444.3       & 1152.4     \\ \hline
8       & 106.0              & 307.4          & 470.2       & 2177.8     \\ \hline

\end{tabular}
\end{table}
\tabcolsep=0.11cm
\begin{table}[]
\caption{Training statistics for different stop and restart configurations}
\centering
\begin{tabular}{|c|c|c|c|c|c|}
\hline
\# $GPUs_{init}$ & $steps_{stop}$ & \#$GPUs_{new}$ & $steps_{tot}$ & $epochs$ & $T_{tot}$ (min) \\ \hline

   1             & $N/A$          & $N/A$          & $62.5k$     & $160$   & $368$ \\ \hline
   2             & $N/A$          & $N/A$          & $33.2k$     & $170$   & $232$ \\ \hline
   4             & $N/A$          & $N/A$          & $15.6k$     & $160$   & $126$ \\ \hline
   8             & $N/A$          & $N/A$          & $8.3k$    & $170$   & $84$  \\ \hline
   4             & $5k$           & $8$            & $10.9k$   & $171$   & $104$ \\ \hline
   4             & $10k$          & $8$            & $12.9k$   & $162$   & $113$ \\ \hline

\end{tabular}
\end{table}

We start by profiling the training runtime of ResNet-110 for the CIFAR-10 dataset. Table 1 confirms our expectation that there should be no statistically significant difference between forward propagation times across various numbers of GPUs. Analysis reveals that backward propagation and \textit{all reduce} operations are performed concurrently, such that gradient exchange occurs concurrently with the computations of other gradients. Overall, the system's scaling efficiency from 4 to 8 GPUs is 94.5\%. \textit{As such, our rescaling algorithm will not suffer significant performance drops from increased number of GPUs for a given job, and associated gains from better resource utilization will remain intact.}

After confirming the scaling efficiency, we experiment with reallocating GPU resources at different points of training. Table 2 shows the different training configurations. The first 3 training jobs are baselines with constant number of resources (GPUs) for the entire training job. We observe the total number of steps and the time it takes to converge with different number of GPUs. We then run experiments with checkpoint, stop, reallocate resources and restart. As shown in the last two experiments in Table 2, we start training with 4 GPUs with minibatch size of 128/GPU. We stop the training at 5k steps (after 51 epochs) and 10k steps (after 102 epochs) marks for experiments 4 and 5 respectively and checkpoint the models. Then we restart the training with 8 GPUs for the training job and continue training from the last checkpoint we saved. When we restart the training, we also readjust the learning rate by a factor of 2 using equation \ref{learning_rate_eqn}. We save close to 50 min ($\sim 32\%$) when we restart the job with 8 GPUs after pausing training at 51 epochs and close to 36 min ($\sim 23\%$) when we pause and restart after 102 epochs. The checkpoint, stop, reallocation, restart times average $\sim 10$ seconds, which is fairly minor in terms of the performance gains by reallocating more resources to the training job. \textit{Consequently, the rescaling algorithm will improve the runtime of jobs by rescaling to take advantage of newly available resources.}

\section{Scheduler simulation}
Using data from the experimental runs we developed a simulation of a scheduler which allocates GPUs to jobs according to a scheduling strategy and predicts the convergence of those runs based on data from previously completed runs.  We run six strategies: A precompute version of the scheduling strategy which assumes minimum data to simulate has been generated by scheduling time,  an exploratory version of the strategy which gives a new job 8 GPUs for the first ten minutes to run for 2.5 minutes at each of 1, 2, 4 and 8 GPUs, and four fixed strategies where each job requests 1, 2, 4 or 8 GPUs respectively.  In the no contention case, each job is able to get it's maximum permitted GPU's under the strategy it's given without conflicts. We simulate job arrival using exponential arrival times resulting in a poisson process.  We use exponential job arrival times of 250s for extreme contention, 500s for moderate contention and 1000s for no contention.  Across all strategies, the extreme contention workload peaks at 125 simultaneous jobs, the moderate workload at 59 simultaneous jobs, and the no contention workload at 20 simultaneous jobs.  The total number of jobs are 206, 114, and 44 respectively. We show results (Table 3) for these arrival times on a simulated 64 GPU cluster.  The eight GPU arrival time is 2.1 times worse than the exploratory scheduling strategy in the moderate contention workload.  The exploratory algorithm is forced to make an explore-optimize tradeoff which works poorly under extreme contention. Similarly, in the zero contention case, the exploration cost underperforms the eight GPU allocation because of the 7.5 minutes spent using less than all eight of the GPUs per job. Regardless, the moderate contention case is consistent with an expected grid workload and in this case both the exploratory and precompute strategies perform well.  When precomputing is feasible, the precompute algorithm always outperforms or ties other strategies.

\begin{table}
\caption {Average job completion times (in hours) for various arrival times and scheduling strategies}
\centering
\begin{tabular}{|c|c|c|c|c|c|}
\hline
Algorithm and Contention & Extreme & Moderate  & None \\ \hline

  Precompute            & \textbf{7.63}          & \textbf{2.63}         & \textbf{1.40}   \\ \hline
  Exploratory             & 20.42         & 2.92          & 1.47     \\ \hline
  Eight            & 22.76          & 6.20          & \textbf{1.40} \\ \hline
  Four            & 12.90           & 3.50            & 2.21  \\ \hline
  Two             & 11.49          & 4.58            & 3.78  \\ \hline
  One             & 10.10          & 6.32            & 6.37  \\ \hline
\end{tabular}
\end{table}

\section{Conclusions} \label{sec:conclusions}

In this paper, we demonstrate the feasibility of better scheduling algorithms for ring architectures, and provide a candidate algorithm based on a new doubling heuristic with good performance in simulation. In particular, we have over double the performance in moderate contention environments. The advances show that communication aware scheduling of ring architecture jobs can improve cluster utilization.

\section{Future Work}
 We would like to expand the range of workloads to a more diverse range of tasks including natural language processing, computer vision, and speech.  We would also like to run the scheduler on a wider range of optimizers.  We plan to have the simulator run real workloads with a simulated schedule rather than using a simulated schedule with simulated workloads. We would like to make further improvements to the explore strategy under extreme contention. Lastly, we would like to be able to setup the scheduler for a real world environment and have it run jobs for a deep learning cluster. 

\bibliographystyle{habbrv}
\bibliography{nips_2018}

\begin{thebibliography}{10}
\expandafter\ifx\csname url\endcsname\relax
  \def\url#1{\texttt{#1}}\fi
\expandafter\ifx\csname doi\endcsname\relax
  \def\doi#1{\burlalt{doi:#1}{http://dx.doi.org/#1}}\fi
\expandafter\ifx\csname urlprefix\endcsname\relax\def\urlprefix{URL }\fi
\expandafter\ifx\csname href\endcsname\relax
  \def\href#1#2{#2}\fi
\expandafter\ifx\csname burlalt\endcsname\relax
  \def\burlalt#1#2{\href{#2}{#1}}\fi

\bibitem{TensorFlow}
M.~Abadi, P.~Barham, J.~Chen, Z.~Chen, A.~Davis, J.~Dean, M.~Devin,
  S.~Ghemawat, G.~Irving, M.~Isard, M.~Kudlur, J.~Levenberg, R.~Monga,
  S.~Moore, D.~G. Murray, B.~Steiner, P.~Tucker, V.~Vasudevan, P.~Warden,
  M.~Wicke, Y.~Yu, and X.~Zheng.
\newblock Tensorflow: A system for large-scale machine learning.
\newblock In {\em 12th USENIX Symposium on Operating Systems Design and
  Implementation (OSDI 16)}, pages 265--283, 2016.
\newblock
  \urlprefix\url{https://www.usenix.org/system/files/conference/osdi16/osdi16-abadi.pdf}.

\bibitem{ImageNet1Hour}
P.~Goyal, P.~Doll{\'{a}}r, R.~B. Girshick, P.~Noordhuis, L.~Wesolowski,
  A.~Kyrola, A.~Tulloch, Y.~Jia, and K.~He.
\newblock Accurate, large minibatch {SGD:} training imagenet in 1 hour.
\newblock {\em CoRR}, abs/1706.02677, 2017,
  \burlalt{1706.02677}{http://arxiv.org/abs/1706.02677}.
\newblock \urlprefix\url{http://arxiv.org/abs/1706.02677}.

\bibitem{Kaiming}
K.~He, X.~Zhang, S.~Ren, and J.~Sun.
\newblock Deep residual learning for image recognition.
\newblock In {\em Computer Vision and Pattern Recognition (CVPR), 2016}, 2016.

\bibitem{ResNetv2}
K.~He, X.~Zhang, S.~Ren, and J.~Sun.
\newblock Identity mappings in deep residual networks.
\newblock {\em ECCV}, 2016,
  \burlalt{1603.05027}{http://arxiv.org/abs/1603.05027}.
\newblock \urlprefix\url{http://arxiv.org/abs/1603.05027}.

\bibitem{CIFAR10}
A.~Krizhevsky.
\newblock Learning multiple layers of features from tiny images.
\newblock {\em University of Toronto}, 05 2012.

\bibitem{NCCL}
Nvidia.
\newblock Nvidia collective communications library v2.
\newblock \url{https://developer.nvidia.com/nccl}, 2017.

\bibitem{OpenMPI}
OpenMPI.
\newblock Open source high performance computing.
\newblock \url{https://www.open-mpi.org/}, 2017.

\bibitem{Optimus}
Y.~Peng, Y.~Bao, Y.~Chen, C.~Wu, and C.~Guo.
\newblock Optimus: an efficient dynamic resource scheduler for deep learning
  clusters.
\newblock In {\em Eurosys 2018, Proceedings of the 13th EuroSys Conference},
  2018.

\bibitem{Rabenseifner}
R.~Rabenseifner.
\newblock Optimization of collective reduction operations.
\newblock {\em International Conference on Computation Science}, 2004.

\bibitem{Horovod}
A.~Sergeev and M.~D. Balso.
\newblock Horovod: fast and easy distributed deep learning in tensorflow.
\newblock {\em CoRR}, abs/1802.05799, 2018,
  \burlalt{1802.05799}{http://arxiv.org/abs/1802.05799}.
\newblock \urlprefix\url{http://arxiv.org/abs/1802.05799}.

\bibitem{MPICH}
R.~Thakur and R.~Rabenseifner.
\newblock Optimization of collective communication operations in mpich, 2005.

\end{thebibliography}

\end{document}